# ОПРЕДЕЛЕНИЕ РАССТОЯНИЯ ДО ОБЪЕКТА В ЗОНЕ ДВИЖЕНИЯ АВТОМОБИЛЯ, ИСПОЛЬЗУЯ АНАЛИЗ ВИДЕОДАННЫХ

Е.В. ЛЕГЧЕКОВА[1], кандидат физико-математических наук, доцент кафедры высшей математики
О.В. ТИТОВ[2], кандидат физико-математических наук, доцент, доцент кафедры естественных наук

[1]*УО «Белорусский торгово-экономический университет потребительской кооперации» г. Гомель, Республика Беларусь*
[2]*УО «Гомельский инженерный институт» МЧС Республики Беларусь, г. Гомель, Республика Беларусь*

Рассматривается метод использования видеокамер, установленных на автомобиле, для вычисления расстояния до объектов в зоне его движения.

**Ключевые слова:** видеокамера, дорожное движение, стереозрение, определение расстояния, система безопасности.

Для снижения числа аварий на дорогах очень важно, что бы водитель мог своевременно среагировать на возникшее на пути автомобиля препятствие. Для этого будет полезно, иметь систему, которая сможет контролировать пространство перед автомобилем, и предупреждать водителя в случае опасности.

Для получения информации целесообразно использовать две камеры, т. к. радарные датчики не могут учесть геометрию дороги (они могут определять только расстояние по прямой). А с помощью камер можно будет контролировать все видимое пространство перед автомобилем, что позволит просматривать и повороты дороги (при условии, что они просматриваются).

Недостатком использования камер является невысокая точность определения расстояния и сложность обработки информации. Следует заметить, что большая точность в данном применении не важна, необходимо просто знать есть препятствие на дороге или нет и примерное расстояние до него в динамике. Для упрощения обработки информации следует использовать новые подходы.

Несмотря на существенные достижения в области распознавания объектов и детектирования их параметров, многие вопросы реализации математических методов в виде программно-аппаратных комплексов являются недостаточно изученными. Это связано в первую очередь с частым представлением общей проблемы в виде двух частей: математической и программно-аппаратной, а при сращивании алгоритмов происходит уменьшение точности в определении параметров целей и увеличение ошибок из-за различных несовместимостей.

Метод определения расстояния по изображениям, полученным с помощью стереозрения, является одним из вариантов определения расстояния до требуемого объекта. Метод предполагает использование двух идентичных камер с определенным расстоянием между ними, которое называется базой.

В случае двух идентичных камер с параллельными оптическими осями расстояние до точки определяется как

$$r_i = \frac{fd}{x_1 - x_2}, \qquad (1)$$

где $f$ – фокусное расстояние; $d$ – расстояние между камерами; $x_1$ и $x_2$ – координаты проекций на левом и правом изображениях.

Для более удобного практического применения формулы (1) представим ее в виде

$$r_i = \frac{dH}{tg\alpha(x_1 - x_2)}, \qquad (2)$$

где $d$ – база (расстояние между камерами); $H$ – горизонтальное разрешение изображения; $\alpha$ – угол обзора камеры; $x_1$ и $x_2$ – координаты точки, до которой определяется расстояние, в координатной системе отсчета первой и второй камеры соответственно.

Для возможности использования формулы (2) считается, что изображения, получаемые с камер, ректифицированы, т. е. камеры расположены так, что в их координатных системах отсчета координаты точки, до которой требуется определить расстояние, $y_1$ и $y_2$ равны, это означает, что горизонтальные линии на изображениях соответствуют одной плоскости.

Трудности использования данного способа заключаются в сложности правильной установки двух камер: оси камер должны быть параллельны друг другу, а также перпендикулярны линии, соединяющей центры камер. Вследствие неправильной установки камер может получиться очень существенная неточность измерения (разница в один градус может привести к погрешности более чем в два раза).

Для уменьшения погрешности предлагается увеличить базу до расстояния того же порядка, что и измеряемое.

Для устранения таких проблем возможно использование в алгоритме методов ректификации изображений [1], но это приводит к серьезному усложнению алгоритма.

Из условия, чтобы чувствительность определения расстояния была высокой (т. е., чтобы изменение разности пикселей на единицу приводило к изменению определяемого расстояния не более чем на 5%), можно определить, начиная с какой разности пикселей следует применять формулу (2):

$$r_1 - r_2 = 0{,}05 r_1;$$
$$\frac{0{,}95 dH}{tg\alpha\, \Delta x_1} = \frac{dH}{tg\alpha\, \Delta x_2};$$
$$\Delta x_1 = 0{,}95 \Delta x_2;$$

Учитывая, что $\Delta x_2 = \Delta x_2 + 1$, получаем $\Delta x_1 = 19$.

Это означает, что, начиная с разности пикселей двух изображений, равной 19, возможно применять метод определения расстояния до объекта с помощью стереовидения.

Для того чтобы можно было определить расстояние вплоть до 500м при угле обзора одной камеры $\alpha = 13^\circ$ и горизонтальном разрешении изображения, равном 1920 пикселей, вычислим по формуле (2) базу $d$:

$$d = \frac{r_i tg\alpha(x_1 - x_2)}{H} = \frac{500 \cdot tg 13^\circ \cdot 19}{1920} = 1{,}14 \text{ м.}$$

На рис. 1 показана зависимость расстояния до исследуемого объекта от разности между изображениями, полученными с двух стереокамер.

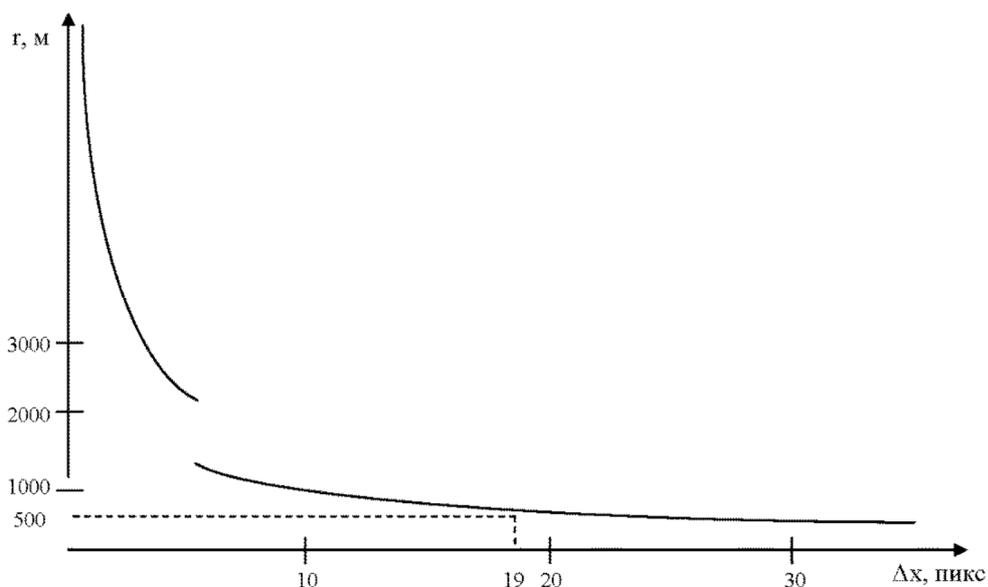

Рис. 1. Зависимость расстояния до объекта от разности пикселей между изображениями объекта с двух камер

Важной задачей определения параметров движущегося объекта является задача определения погрешности измерения и интервалов применимости предлагаемой методики. На рис. 2 представлена зависимость ошибки определения положения объекта от погрешности определения угла между камерами стереопары при различных величинах базы (0,60 м, 1,00 м, 1,14 м). Объект находится на расстоянии 500 м. Ошибка в один градус приводит к неточности в 150–300 %.

Погрешность данного метода будет зависеть от: погрешности нахождения первоначального расстояния, найденного при помощи пропорции, погрешности способа стереозрения (чувствительности способа – чем ближе будет объект, тем меньшее изменение расстояния происходит при изменении разности пикселей изображений на единицу), а также от погрешности метода распознавания объекта, то есть возможности точного выделения его границ.

Зависимость ошибки измерения расстояния методом стереозрения от расстояния до объекта:

$$\varepsilon = \frac{r_{i+1} - r_i}{r_{i+1}}. \qquad (4)$$

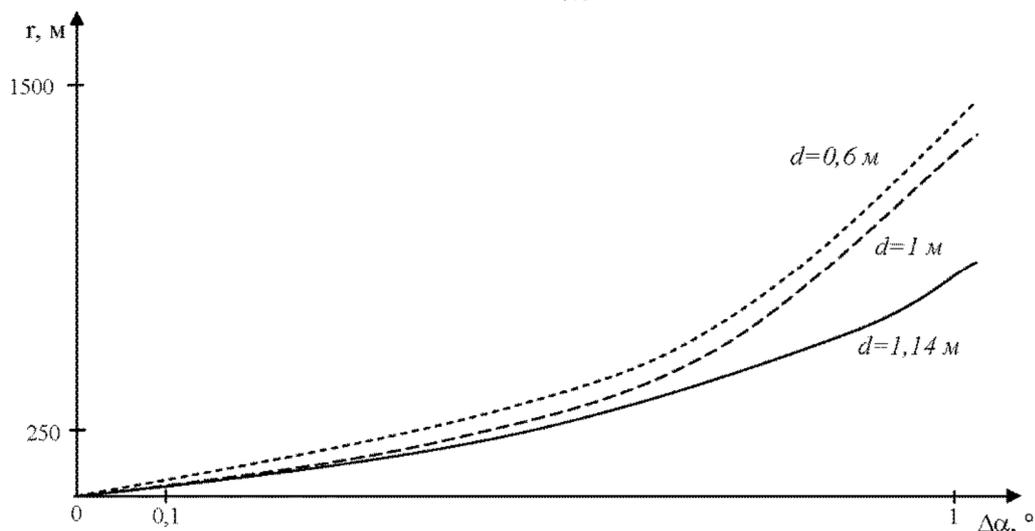

Рис. 2. Зависимость погрешности определения положения объекта от ошибки определения угла между осями камер (при величине стереобазы 0,60 м, 1,00 м, 1,14 м)

Ниже приведен график зависимости погрешности метода стереозрения от расстояния до объекта для различных геометрических размеров изучаемой цели.

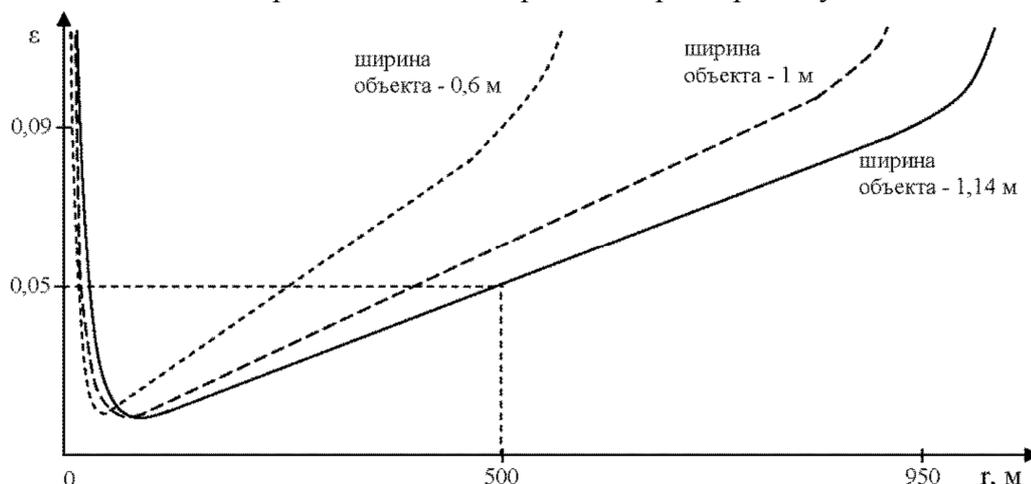

Рис. 3. Зависимость погрешности метода стереозрения от расстояния до объекта для различных геометрических размеров изучаемой цели

Результаты математического моделирования процесса стереораспознавания позволили спрогнозировать расстояние до определяемого объекта по количеству различающихся пикселей с двух синхронных изображений, вычислить зависимость точности определения положения изучаемого объекта от первоначальной ошибки во взаимном ориентировании камер.

В дальнейшем предполагается усовершенствовать математический и программный аппарат для увеличения точности детектирования объекта и возможности использования спроектированного комплекса в реальных задачах обеспечения безопасности и соблюдения ПДД на автомобильных дорогах.

В самом простом варианте данные с камер могут выводиться водителю (на панель приборов) в виде расстояния до впереди идущего автомобиля. При быстром сближении можно выдавать предупреждающий сигнал. Полученные в результате обработки данные могут быть использованы для комплексной системы безопасности, которая сможет не только предупредить водителя но и самостоятельно остановить автомобиль в случае необходимости.

Описанные в данной статье методы обработки данных с камер позволяет получить дешёвый датчик с широким спектром применения, для автомобильных систем безопасности.

**E.V. Legchekova, O.V. Titov**
**CALCULATE DISTANCE TO OBJECT IN THE AREA WHERE CAR, USING VIDEO ANALYSIS**

The method of using video cameras installed on the car, to calculate the distance to the object in its area of movement.